# Multi-path Convolutional Neural Networks for Complex Image Classification

Mingming Wang
Dalhousie University
m.wang@dal.ca

**Abstract**

*Convolutional Neural Networks demonstrate high performance on ImageNet Large-Scale Visual Recognition Challenges contest. Nevertheless, the published results only show the overall performance for all image classes. There is no further analysis why certain images get worse results and how they could be improved. In this paper, we provide deep performance analysis based on different types of images and point out the weaknesses of convolutional neural networks through experiment. We design a novel multiple paths convolutional neural network, which feeds different versions of images into separated paths to learn more comprehensive features. This model has better presentation for image than the traditional single path model. We acquire better classification results on complex validation set on both top 1 and top 5 scores than the best ILSVRC 2013 classification model.*

## 1. Introduction

Convolutional Neural Networks [6],[10],[11],[12],[20] show significant advantages on image classification tasks. Especially, in recent ImageNet Large-Scale Visual Recognition Challenges [18], series models make great performance improvement, such as [9] achieved top 5 error rate at 16.4% in 2012 and [21] reached top 5 error rate at 14.7% in 2013. There are several factors that fasten the progress made on image classification by convolutional neural networks, such as the emergence of large scale labeled image datasets and high performance computing chips, especially modern GPU with thousands of stream processors, which make training a bigger model with millions of parameters in parallel become feasible.

Although great progress has been made, there is still very limited performance analysis in corresponding with internal operations and how image data impact performance. More specifically, what images get good results and what images do not perform so well? Why do some images have performance drop and how can we improve it? We firstly design an experiment to compare the performance of a single path convolutional neural network model on 2 groups of datasets that have different image complexity. Both groups contain training and validation images with equal quantity and class labels. The first group contains the simplest images and the second group contains the most complex images. We run the model on 2 groups of datasets independently and then provide thorough results analysis combined with learning curves. After we go through the internal steps of the model, we clearly point out the weaknesses of convolution and pooling operations, which are more likely extract high frequency components of images. This mechanism causes simple shape and less textured objects gradually disappear in clustered background after several iterations of convolution and pooling. Therefore, the final feature vector cannot present foreground objects well and then gets bad result for prediction. We define a novel multiple paths network to overcome the weakness of single path structure network. These paths learn different aspects of images and retain more valid info than single path. The result proves our method is more accurate than ILSVRC 2013 best model [21] for classification on complex image datasets.

### 1.1 Related work

Convolutional Neural Networks with millions of parameters are by far the largest model for images classification task. Researchers often feed the network with large datasets and train for days or weeks in order to make sure network converge. Thus, there are many methods try to speed up training procedure. The common method is to adopt parallel computing technology, especially under multiple GPU or GPU hardware configuration, because feature learning of each layer is similar and easy to scale. There are 2 major ways for parallel training: model parallelism and data parallelism [8]. Model parallelism is training different parts of the network on different hardware, such as training the first layer on GPU 1 and training the second layer on GPU 2. These 2 layers need to synchronize with each other. Data parallelism divides big datasets into smaller ones and trains on different GPUs at same time. The goal of parallel training method is shorten training time. However, this approach cannot improve classification accuracy and overcome the weakness of single path convolutional neural networks. As shown in [8], this model uses 8 GPUs is not better than 1 GPU for classification result. They both have 42% top 1 error rate over 1000 image classes. Furthermore, the computation and network complexity increase on the parallel method, since the multiple GPU model uses more complicated feed forward and backward strategy and needs synchronize the weights between each path.

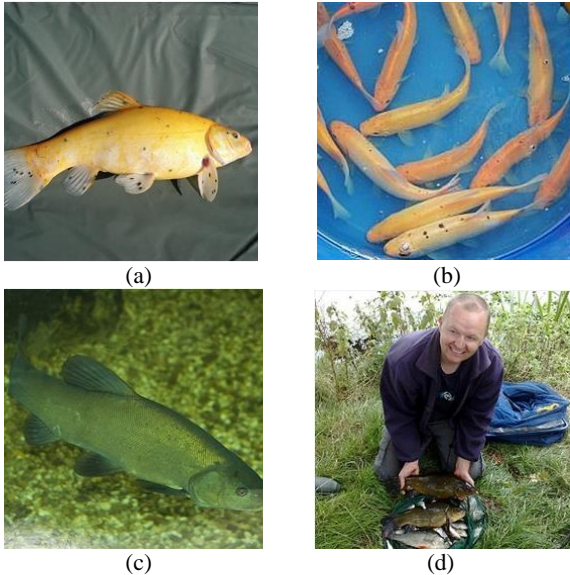

**Figure 1**. Four images from ImageNet [7] dataset of same class "Tench, Tinca tinca". From (a) to (d) images become more complex

Some other parallel methods like [3] use different resolution images which feed into different paths. After several convolution and pooling steps, feature maps from final stage are aligned into same size by up scaling smaller ones. This method tries to learn features of different pyramid level of image that essentially does not speed up the learning speed. Model [1] uses multiple columns to train images, with each column training a processed block in different way. The final prediction is averaged over multiple columns.

## 2. Dataset

ImageNet [7] Dataset for classification task has 1.27 million training images and 50,000 validation images. They were collected from the internet and manually labeled by using Amazon Mechanical Turk crowd sourcing tool. These images spread over 1000 classes, for example natural scenes, animals and human made objects. This dataset is by far the most difficult challenge for image classification algorithm. The reasons are not only the large number of images and high varieties within each class, such as different background, resolution, objects combination, exposure and angles, but also the similarities between different classes. For example, there are more than 100 species of dogs. They look alike but have different class labels. Other animal for instance "kit fox" is included in this data set either. It has similar shape and fur pattern with dog. Some image classes belong to human made objects like "beer glass", "beer bottle", "wine bottle" and "cocktail shaker". They usually have same background setting and mix together. "Beer glass" appears with "beer bottles"; "wine bottle" appears with "beer glass"; "cocktail shaker" combines with other bottles and glasses. Human made objects have similar size and shape. Many of them have a simple shape and less texture, which make these objects not easy to be distinguished between each other, especially the foreground objects which only occupy a small area of the image or are far away from camera. Another kind of image class is human made scene, for instance "toy shop". The images in this class do not share common objects. They could be any toys or outdoor picture of toy shop only. All above conditions dramatically increase the difficulty for training and prediction. ILSVRC [18] classification contest uses top 5 error rate for model evaluation. If the true validation image label is among top-5 model predict labels, then the result is marked as right, or else the prediction is wrong.

To save training time, we choose 100 of 1000 categories at step 10 based on ImageNet categories list. This subset includes many animals and 14 of 100 categories are different species of dogs, for example "silky terrier", "white terrier", "cairn terrier", "cocker spaniel" and "Scottish deerhound", etc. Human made objects classes contain "beer glass", "wine bottle", "cocktail shaker", "pill bottle", etc. Human made scenes include "toy shop". Most of classes have 1300 training images. If some classes have less than 1300 images, we create new training images by randomly choosing from existing ones, then flip right and left to make these classes have just 1300 images. As all image data are collected from the internet, they have different sizes, color channels in RGB, CMYK and gray. We transform all images in both training and validation sets into 3 channels RGB format, resize and crop into same size at $256 \times 256$. All images are subtracted into zero mean instances before they are fed into the network.

### 2.1 Image complexity analysis

Most of popular convolutional neural network models train source images minus global mean value directly. Image processing is not involved much. We do more investigation of image data to see how different images impact the learning curve of the model. Now we have a subset of ImageNet dataset with 100 classes and each class has 1300 training images. After we go through these images, we find big varieties within each class. As shown in figure 1, these four images are from the same class "Tench, Tinca tinca". It is a freshwater dace-like game fish of Europe and western Asia noted for ability to survive outside water. Image (a) only has one fish and the background is flat. Image (b) has fifteen fish and background is still plain. Image (c) has one fish but with a much textured small cobblestone background. It is visually more complex than previous 2 images. The last image (d) is the most complex one. It shows many different objects with clustered grassland background, including a human being. Fish only occupy a very small region of the whole image in the bottom. However, this image still has the same label as all three images. We suppose (a) and (b) are good training cases; (c) and (d) are not ideal training cases. Convolutional neural networks can extract more features from foreground fish in the first two images. In comparison, the last two, especially (d) is too noisy. It contains more than one object, but not limited to fish. So the final feature vector includes the abstraction of multiple objects.

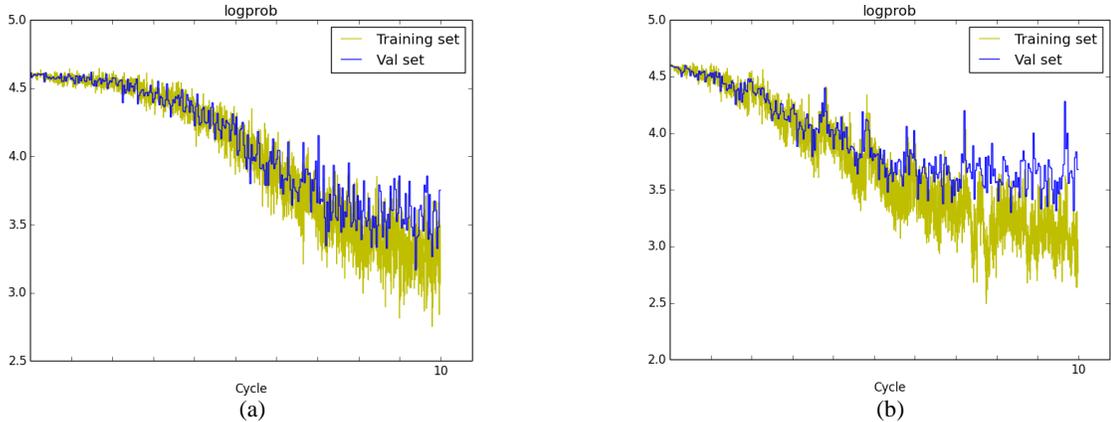

**Figure 2**. Learning curves for 2 groups of training and validation images. (a) Shows learning curve for group 1 with simple images. (b) Shows learning curve for group 4 with complex images.

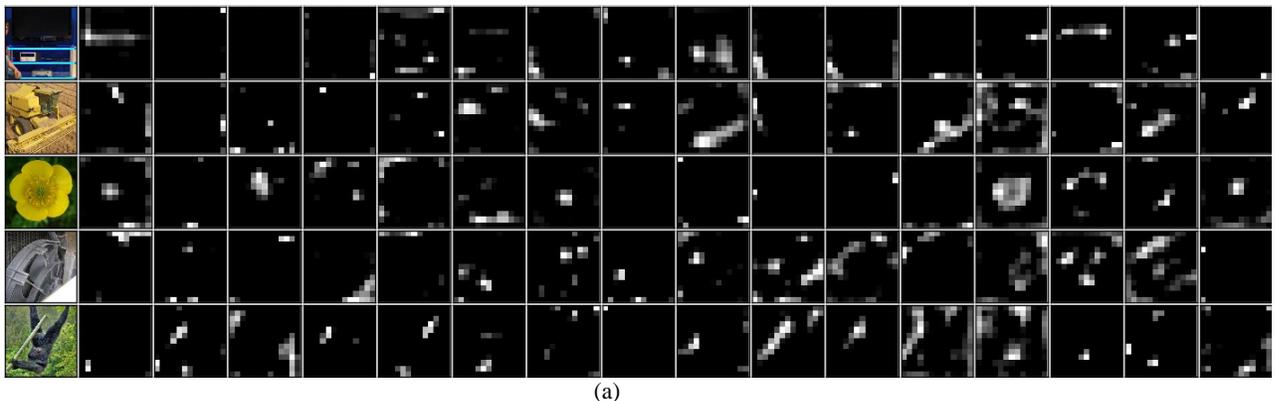

(a)

**Figure 3**. Random subset feature maps of the last convolutional layer from [9].

Overall, we presume simple images of each class are ideal training examples and complex images are worse examples. Complex images are more likely to contain clustered background and irrelevant objects. There are many ways to measure image complexity, such as entropy and wavelet [2],[13] coefficient. We use 2D wavelet transform, as it computes very fast and is sensitive to different orientation. Large gradient in local image region has large wavelet coefficient. Image complexity has positive correlation with the quantity of large coefficients. We transform all 256×256 training and validation RGB images into gray scale, perform first order wavelet transform and normalize wavelet coefficients between 0 and 1. We define an indices $C$ that counts the number of normalized coefficient $d_{x,y}^k$ which is larger than 0.5 threshold from three detail images in horizontal, vertical, and diagonal directions, where $k$ is index of detail images and $x, y$ are pixel coordination. This $C$ indices measures the complexity of each image.

$$C = \sum_k \sum_{x,y} c_{x,y}^k \begin{cases} 1, & d_{x,y}^k > 0.5 \\ 0, & d_{x,y}^k \leq 0.5 \end{cases}$$

We sort all 1,300 training images of each class by this indices and separate them into 4 groups. The first group has 325 images with small indices. They are simple and ideal training examples. The fourth group contains the most complex images with large indices. This approach is applied to the validation set as well. Each class of validation set has 50 images which is not divisible by 4. So we choose the top 48 images after sorted by indices $C$ and then separate them into 4 groups. Each group has 12 validation images for each class. The first group contains simple validation images and the last group is the most complex validation images. We pair the *i*th group training image set with the *i*th group of the validation image set.

## 3. Compare simple and complex images on network

We reproduce [21] model network structure definition and parameters setting. It has the best result for the image classification task on ILSVRC 2013. This single path model has 5 convolution layers and 2 full connection layers. To make training faster, we do not use dropout [5] for full connection layers. The last full connection layer connects to the softmax layer for classification.

We run group 1 simple image dataset and group 4 complex image dataset on above model independently for 10 cycles. Each data dictionary of training and validation set contains 100 images with 100 different class labels. Image sequence in each data dictionary is randomly permuted. Validation frequency is set to 10, which means it runs 1 validation data dictionary after running 10 training data dictionaries. Both groups have 325 training images for each class and total class number is 100. So we have 32,500 training images for both groups. The validation set of each group has 1,200 images. We adopt data augment method in [9] to make more training cases through image translation and horizontal reflections. This is done by randomly cropping 224×224 region of 256×256 source image and flipping in horizontal direction at 50% chance. For validation set data, we crop center 224×224 region of image. As we run training set for 10 cycles, so the valid training cases are up to 325,000 after applying the data augment method.

The learning curves are shown in Figure 2. Obviously, group 1 with simple images has a smoother learning curve in (a). For this group, the validation set curve in the blue line has less over fitting, since it does not detach from the training curve very much. The log probability of validation set after 10 cycles is near 3.5. For group 4 with complex images, both training and validation curves fluctuate a lot. Training and validation curve detach very quickly. This is a very typical over fitting phenomenon. The parameters learned from complex training images are not fit for complex validation images well. It can be explained in another aspect, complex images within each class are not similar with each other or have big variance. They could share the same foreground object, but the background varies a lot. Therefore the final feature vectors have big gap. We can make conclusion that for convolutional neural networks, complex images are more difficult to learn.

In fact, many other image classification algorithms have similar issues. They do not perform well with complex images. We need find out the reasons from internal steps of convolutional neural networks that make performance drop on complex images.

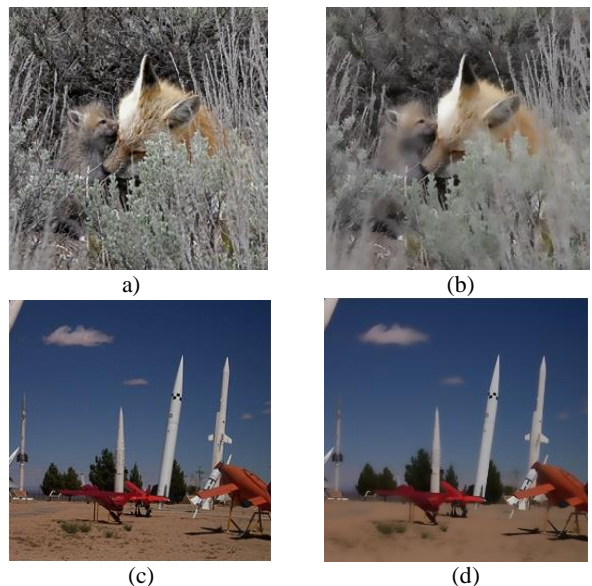

**Figure 4**. (a) and (c) from ImageNet [7] dataset. (b) is bilateral filtered image of "kit fox" under clustered background. (d) is bilateral filtered image "missile".

### 3.1 Performance drop analysis

Let's review convolution and pooling process to see possible issues of the model. The whole training process is trying to learn weights, bias and filters through network forward and backward. The filters in the first convolutional layer learn the basic pattern of local region of image, such as strong colors and high frequency components edges and texture. The feature maps are outputs of convolution operation between learned filters and source image and then pass through activation function like nonlinearity ReLU [14] below.

$$f(x) = \max(0, x)$$

This nonlinearity function maps negative value and zero to zero and retain all positive values directly. If filters have high similarity with local region of image, the feature map will get a large activation value. Edges like filters take very large proportion for layer 1, which mainly contains high frequency components. Therefore, high frequency regions in source images are more likely to be retained in feature maps. Pooling is a subsampling process. It chooses the max value of sub region of feature map. This operation enhances the high frequency components which are retained for the following steps. After several rounds of convolution and pooling, low frequency components that correspond with plain regions of source image are dropped gradually. As shown in figure 3, the left column is source images and following images on the right are random feature maps of the last convolutional layer. By comparing feature maps and source image, we can see that only high

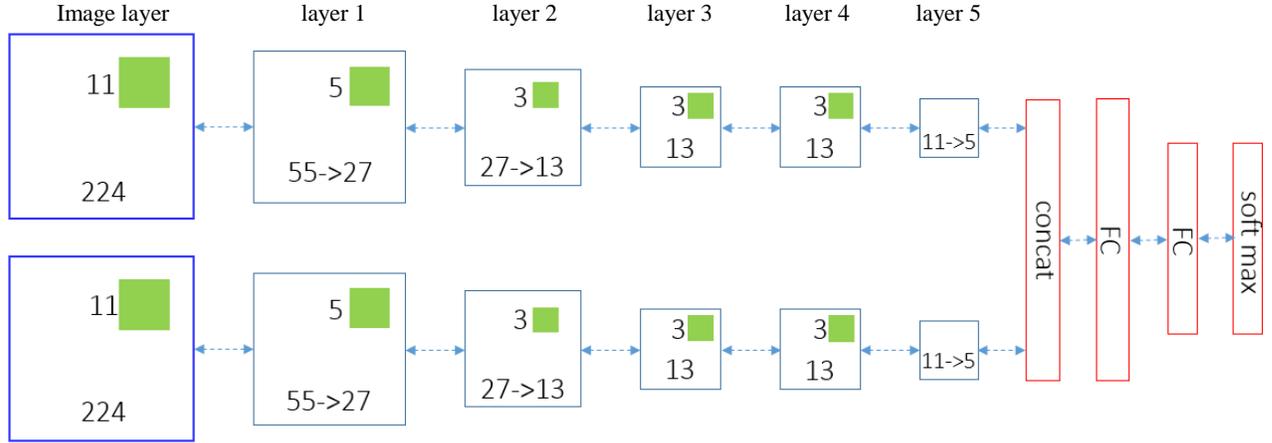

**Figure 5**. Multi-path convolutional neural network architecture. There are 2 paths. The first path feed with source image and the second path feed in bilateral filtered image. 2 paths concatenate afterwards. We use 2 full connection layers then connect to softmax regression layer for output

frequency regions, like edges and textures area have activation values in feature maps and all rest areas are zeros.

We need to pay special attention to the last row of figure 3. The source image on the left has very clustered background. As feature maps show on the right, most of activation values are corresponding to high frequency grass background and border of foreground. However, the area inside of foreground does not have any activation values in any feature maps at all. This leads to foreground object is almost eliminated in the final feature vector. If feature vector does not contain or only contains very limited info for foreground object, then classification performance must decrease. The goal of image classification algorithms, including convolutional neural network is to extract valid features from images. Nevertheless, based on our observation and analysis, convolutional neural network is more sensitive to high frequency components of images. The real situation is not all image objects in foreground are very complex and highly textured. Many human made objects usually have simple shapes and less texture. If these simple objects are set in complex background, they could be gradually eliminated during convolution and pooling process. [18] indirectly proves our conclusion. As shown in this paper, the easiest classes for classification task are "tiger", "hen-of-wood", "porcupine", etc. They all have very unique shape with rich texture. The hardest classes are "ladle lope" and "letter opener", etc. They are very simple objects and lack of texture. Once these simple objects are in complex background, they fail to compete with complex background during convolution and pooling. As a result, the classification performance is bad.

### 3.2 Suppressing high frequency components

Now we find out the root cause why convolutional neural network does not perform well on complex images for each class. The next step is to find a solution to overcome this weakness. As we know the high frequency regions of image are more likely clustered background or irrelevant objects, so we need to suppress high frequency components. There are many valid methods, like the 2D Gaussian blur filter. But this filter removes details of images not only from textured area but also edges. For image classification task, the edges of foreground object are critical info. If we apply Gaussian filter directly, convolutional neural networks cannot learn distinguish features from blur images. Hence, we need to find another method which reduce high frequency components but keep edges at same time. Many filters can smooth image and preserve edge, like Bilateral filter [16],[17],[19] and derived one such as Guided Image filter [4]. Bilateral filter has fast implementation like [15]. The main approach is to compute each pixel by weighted intensity values from nearby pixels. The weights depend on both spatial domain, and radiometric, such as color intensity, depth distance, etc. This preserves edges by systematically looping through each pixel and adjusting weights to the adjacent pixels accordingly. We set half kernel size to 5, spatial domain standard deviation to 3 and radiometric standard deviation to 0.15. These parameters make sure high frequency components are suppressed but the whole image is not too cartoonlike. Blur filtering process reduce high frequency signal much faster than low frequency signal. Figure 4 shown the examples of source images and bilateral filtered images. The foreground "kit fox" in (a) has some blur effect after filtering in (b), but the overall shape and contour are still clear. In contrary, background high frequency details reduce a lot. The background grass of "missile" in (c) has blur effect after filtering in (d), but foreground missiles almost have no change. Suppressing high frequency components of source images makes final feature vector contains less cluster background, then the proportion of foreground appear in feature vector is increased. This means more valid features are extracted

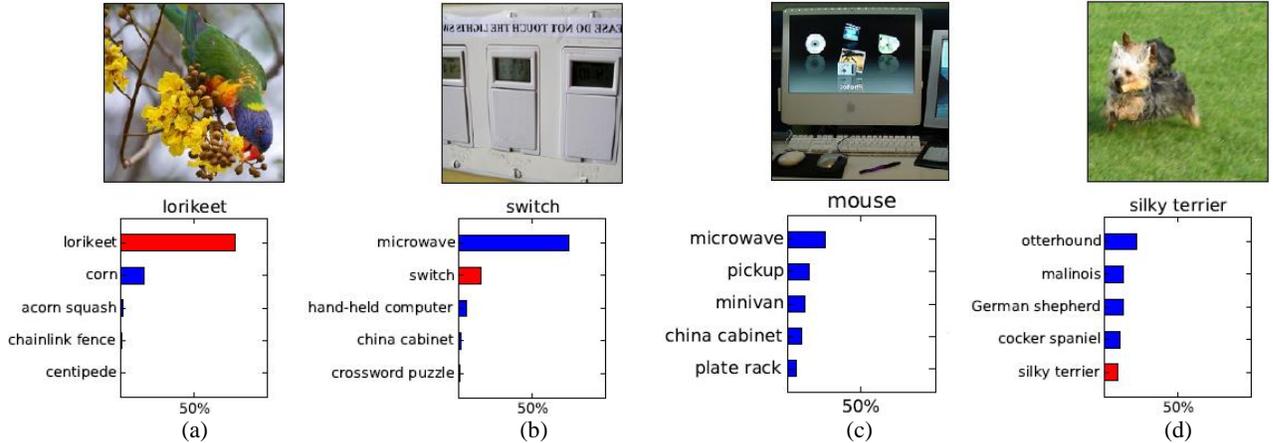

**Figure 6**. Classification result of 4 images in group 4 validation set from ImageNet [7] dataset. Top five predicted labels are most probable ones. Correct label is under each image and the probability of correct label is in red. (a) hits top 1, (b) and (d) hit within top 5. (c) gets wrong label. The mouse in this image is too small.

from the network.

## 4. Multi-path network design and experiment

We define a model with multi-path network as show in figure 5. It extracts features from both source images and bilateral filtered images. The first path feeds source image and the second path feeds bilateral filtered images. This method can learn more comprehensive features than single path. Multi-path model constructs a longer feature vector based on concat layer. It contains features from both paths that better represent the image. If the foreground object is simple and background is complex, then foreground object is more likely retained in the second path, which feeds bilateral filtered image. By contrast, a complex foreground object like "tiger" is more likely retained in the first path, which is dominated by high frequency components. Multi-path model can capture a broader range of foreground objects from simple to complex. Thus, it is more robust than single path.

Detailed network design as follow. Image layer provides 224×224×3 RGB image. In layer 1 to 5, cross feature maps normalization is applied after convolution. The first convolution layer apply 11×11 filter with image layer and output 55×55 feature map. Pooling produces 27×27 output from normalized feature maps. The second convolution Layer applies 5×5 filter with 2 pixels border padding for input feature map. So it produces 27×27 feature maps. The following pooling operation produce 13×13 output for normalized feature maps. Layer 3 and 4 use 1 pixel border padding for input feature maps before convolution. These 2 layers do not use pooling operation anymore. Layer 5 applies 3×3 filter with normalized feature map from output of layer 4. The last pooling operation for layer 5 produce 5×5 output. The numbers of filters in layer 1 to 5 are 48, 192, 256, 256 and 192.   So the length of concat layer is 9600 that combines the output of pooling operation of layer

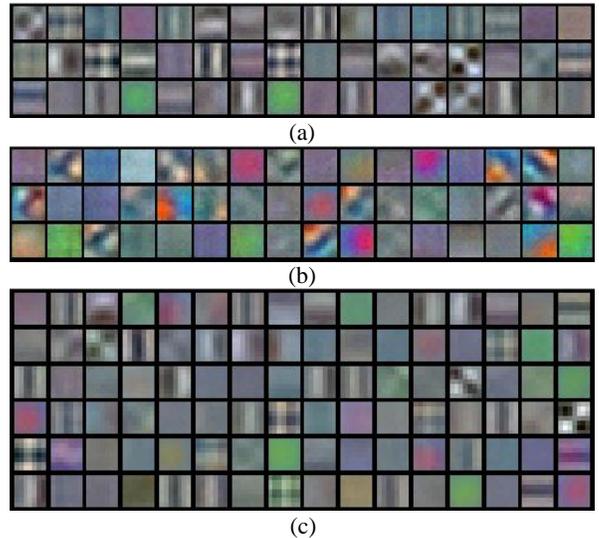

**Figure 7.** (a) and (b) show total 96 filters of first and second paths of our model. (c) shows 96 fitlers learning of [21] single path model.

5 from 2 paths, with each of path contributing 4800 (192×5×5) elements. This feature vector is longer than traditional 1 path model, like 4096 elements in [21]. The following 2 layers are full connection layers. The first full connection layer has the same length as concat layer and the second one with length 100. We leverage dropout [5] layer for 2 full connection layers for input connections. Dropout rate is 0.5. The output of second full connection layer is connected to 100 way softmax which produces distribution for 100 classes. We set all initial value of weights for all layers through zero mean normal distribution with standard deviation 0.01. The initial value of biases with constant 0. The model maximizes multinomial logistic regression function, which maximize the average log-probability of image labels over prediction distribution for all training examples.

We feed the model with the group 4 dataset which has the most complex images for training and testing. This is the most challenging group than other 3 groups. We use the data augment method as described in section 3. Our code running under Windows 7 64bit OS on single NVIDIA 4GB GTX 760 GPU. It costs 8 days for 20 cycles training and testing. We reduce the learning rate by multiply 0.1 after error rate on validation curve goes flat.

## 5. Result

For models comparison, we run the group 4 dataset on [21] single path model (with dropout layer) for 20 cycles as well. They both use 1,200 center crop 224×224 validation images to report score. The results are shown in table 1. We win in both top-1 and top-5 results.

| Model | Top 1 | Top 5 |
|---|---|---|
| ILSVRC 2013 [21] | 66.5% | 35.6% |
| Our model | 64.2% | 33.5% |

**Table 1**, models comparison on the most complex 1,200 validation images. Report top 1 and top 5 error rate

We show the filters of our model from layer 1 of each path in Figure 7. The first 48 filters in (a) from the first path shows very clear patterns, such as vertical and horizontal edges and textures. Most of filters in (a) contains high frequency components. The second 48 filters in (b) from the second path shows more strong color spots but less edges. This is caused when images in the second path are bilateral filtered, which reduce high frequency components. The total 96 filters in 2 paths learn different aspects of images. They are richer than 96 filters learned only from traditional 1 path as show in (c) of model [21]. Furthermore, in our model the filters between 2 paths have no correlations in (a) and (b), but filters within each path show more correlations and similarities. (a) and (c) have similarities, because they both learn from source images.

Figure 6 shows four validation images predictions of our model. Image (a) hit top 1 accuracy. The second probable label "corn" is caused by yellow flowers, which has similarities with "corn" class. Image (b) belongs to a human make object, which has simple shape and less texture. The contour of "microwave oven" look likes "switch" that taken from very near from camera. It hits within top 5. The image (d) is "silky terrier". It is a complex grassland background. The top 5 prediction labels for this image are all dogs, such as "otter hound", "malinois", "german shepherd", "cocker spaniel" and "silky terrier". These 5 classes share high similarities between each other. Thus prediction probable values have no big differences as show in blue bars. Image (c) does not get the right answer because the mouse in this image is too small.

### 5.1 Model generalization

We do a further test to see the model generalization ability. Our model uses group 4 with complex images for training. Now we test the model by using the simplest validation images of group 1. We apply this test method to model [21] to compare performance. They both use 1,200 center crop 224×224 validation images of group 1 to report score. The results are shown in table 2. We win in both top-1 and top-5 results again. Top-1 result has 4.4% improvement.

| Model | Top 1 | Top 5 |
|---|---|---|
| ILSVRC 2013 [21] | 67.6% | 36.6% |
| Our model | 63.2% | 34.9% |

**Table 2**, models comparison on the simplest 1,200 validation images. Report top 1 and top 5 error rate

## 6. Conclusion

In this paper, we do advanced research of convolutional neural network with image analysis and find out the root cause of performance drop on complex images. To resolve the issue, we leverage image processing technology to suppress high frequency components but keep overall object shape and contour. We design and implement multi-path networks that extract more distinguishable features from both source images and bilateral filtered images than single path model that feed source images only. We get more competitive performance improvement on both top 1 and top 5 error rates. Therefore, we can conclude that multi-path convolutional neural network show more advantages than simple path network definition for complex images.

We can make further exploration. For example add in more paths and different path use different depth and parameters setting. We suppose simple and complex objects need different abstraction level. Simple objects should use less convolution and pooling steps. We would also try multiple GPU configuration, which let different path run different GPU to speed up training procedure.